\documentclass{article}
\usepackage{ijcai18}
\usepackage{booktabs}
\newcommand*{\rom}[1]{\expandafter\@slowromancap\romannumeral #1@}
\newcommand\Tau{\mathcal{T}}
\newcommand\Alpha{\mathcal{A}}
\newcommand\Mu{\mathcal{M}}

\newcommand\I{\mathcal{I}}
\usepackage{algorithm}
\usepackage{algpseudocode}
\usepackage{subcaption}
\usepackage{array}
\usepackage{mathtools}
\usepackage{tikz}
\usepackage{scalefnt}
\usepackage{array}

\newcolumntype{M}[1]{>{\centering\arraybackslash}m{#1}}
\newcolumntype{N}{@{}m{0pt}@{}}

\usepackage{amsfonts}
\title{Equations}

\usepackage{times}
\usepackage{xcolor}
\usepackage{soul}
\usepackage[utf8]{inputenc}
\usepackage[font=small]{caption}

\usepackage{fancyhdr}

\title{Behavioral Cloning from Observation}

\author{
Faraz Torabi$^1$, 
Garrett Warnell$^2$, 
Peter Stone$^1$
\\ 
$^1$ The University of Texas at Austin \\
$^2$ U.S. Army Research Laboratory\\
\{faraztrb,pstone\}@cs.utexas.edu,
garrett.a.warnell.civ@mail.mil
}
\begin{document}
	\thispagestyle{fancy}
	\maketitle
	
	\begin{abstract}
		Humans often learn how to perform tasks via imitation: they observe others perform a task, and then very quickly infer the appropriate actions to take based on their observations.
		While extending this paradigm to autonomous agents is a well-studied problem in general, there are two particular aspects that have largely been overlooked: (1) that the learning is done from observation {\em only} (i.e., without explicit action information), and (2) that the learning is typically done very quickly.
		In this work, we propose a two-phase, autonomous imitation learning technique called {\em behavioral cloning from observation (BCO)}, that aims to provide improved performance with respect to both of these aspects.
		First, we allow the agent to acquire experience in a self-supervised fashion.
		This experience is used to develop a model which is then utilized to learn a particular task by observing an expert perform that task without the knowledge of the specific actions taken.
		We experimentally compare BCO to imitation learning methods, including the state-of-the-art, generative adversarial imitation learning (GAIL) technique, and we show comparable task performance in several different simulation domains while exhibiting increased learning speed after expert trajectories become available.
	\end{abstract}
	
	\section{Introduction}\label{introduction}

	The ability to learn through experience is a hallmark of intelligence.
	Humans most certainly learn this way, and, using {\em reinforcement learning (RL)} \cite{sutton1998reinforcement}, autonomous agents may do so as well.
	However, learning to perform a task based solely on one's own experience can be very difficult and slow.
	Humans are often able to learn much faster by observing and imitating the behavior of others.
	Enabling this same ability in autonomous agents, referred to as {\em learning from demonstration (LfD)}, has been given a great deal of attention in the research community \cite{schaal1997learning,argall2009survey}.
	
	While much of LfD research is motivated by the way humans learn from observing others, it has largely overlooked the integration of two important aspects of that paradigm.
	First, unlike the classical LfD setting, humans do not typically have knowledge of the precise {\em actions} executed by demonstrators, i.e., the internal control signals demonstrators use to guide their behavior
	Second, humans are able to perform imitation without needing to spend a lot of time interacting with their environment after the demonstration has been provided.
	
	Most LfD work is unlike human imitation in its assumption that imitators know the actions executed by demonstrators.
	Human imitators typically do not have access to this information, and requiring it immediately precludes using a large amount of demonstration data where action sequences are not given.
	For example, there is a great number of tutorial videos on YouTube that only provide the observer knowledge of the demonstrator's state trajectory.
	It would be immensely beneficial if we could devise LfD algorithms to make use of such information.

\begin{figure}
\begin{center}
\begin{tikzpicture}

\draw [fill={rgb:blue,1;red,1;white,7}, rounded corners] (1.7,0) rectangle (7.9,4);

\node [text={rgb:blue,3;red,1}, align=center] at (4.75,3.7) {\scalefont{0.8} Behavioral Cloning from Observation (BCO)};

\node [align=center] at (0.5,3.9) {\footnotesize START};

\draw [ultra thick,->] (0.5,3.7) -- (0.5,3.2);

\draw [rounded corners](-0.6,2.2) rectangle (1.6,3.2) node [pos=0.5, align=center] {\footnotesize Initialize\\ \footnotesize policy $\pi_\phi^{i=0}$};

\draw [ultra thick,->] (1.6,2.7) -- (2.5,2.7);
\node [align=center] at (2.1,2.95) {\footnotesize $\pi_\phi^{0}$};

\draw [rounded corners](-0.6,0.1) rectangle (1.6,1.1) node [pos=0.5, align=center] {\footnotesize State-only\\ \footnotesize demonstrations};

\node [align=center] at (3.4,0.2) {\footnotesize $D_{demo}$};
\draw [ultra thick,->] (1.6,0.4) -- (4.6,0.4) -- (4.6,0.6);

\draw [rounded corners, fill=white] (2.5,2.2) rectangle (3.9, 3.2) node [pos =0.5, align=center] {\footnotesize Run\\ \footnotesize policy $\pi^i_\phi$};

\node [align=center] at (4.85,2.95) {\footnotesize $\{(s^a_t, s^a_{t+1})\}$};
\node [align=center] at (4.85,2.45) {\footnotesize $\{a_t\}$};
\draw [ultra thick,->] (3.9,2.7) -- (6,2.7);

\draw [rounded corners, fill=white] (6,2.2) rectangle (7.6, 3.2) node [pos =0.5, align=center] {\footnotesize Append to\\ \footnotesize $\Tau_\pi^a, \Alpha_\pi$};

\node [align=center] at (7.35,1.95) {\footnotesize $\Tau_\pi^a, \Alpha_\pi$};
\draw [ultra thick,->] (6.8,2.2) -- (6.8,1.6); 

\draw [rounded corners, fill=white] (6,0.6) rectangle (7.6, 1.6) node [pos =0.5, align=center] {\footnotesize Update\\ \footnotesize model $\Mu_\theta^i$};

\draw [ultra thick,->] (6,1.1) -- (5.2,1.1);
\node [align=center] at (5.65,1.35) {\footnotesize $\Mu_\theta^i$};

\draw [rounded corners, fill=white] (4.2,0.6) rectangle (5.2, 1.6) node [pos =0.5, align=center] {\footnotesize Infer\\ \footnotesize actions};

\draw [ultra thick,->] (4.2,1.1) -- (3.2,1.1);
\node [align=center] at (3.75,1.35) {\footnotesize $\tilde{\Alpha}_{demo}$};
\node [align=center] at (3.75,0.85) {\footnotesize $S_{demo}$};

\draw [rounded corners, fill=white] (1.8,0.6) rectangle (3.2, 1.6) node [pos =0.5, align=center] {\footnotesize Update\\ \footnotesize policy $\pi^i_{\phi}$};

\draw [ultra thick,->] (2.5,1.6) -- (2.5,2) -- (3.5,2) -- (3.5,2.2);
\node [align=center] at (2.2,1.9) {\footnotesize $\pi_\phi^i$};

\end{tikzpicture}
\caption{Behavioral Cloning from Observation (BCO($\alpha$)) framework proposed in this paper. The agent is initialized with a (random) policy which interacts with the environment and collects data to to learn its own agent-specific inverse dynamics model.
	Then, given state-only demonstration information, the agent uses this learned model to infer the expert's missing action information.
	Once these actions have been inferred, the agent performs imitation learning. The updated policy is then used to collect data and this process repeats.}
			\label{fig:framework}
\end{center}
\end{figure}

	Another challenge faced by LfD is the necessity of environment interaction, which can be expensive in several regards.
	One is the amount of time it requires: executing actions, either in the real world or in simulation, takes time.
	If a learning algorithm requires that a large number of actions must be executed in order to find a good imitation policy after a demonstration is presented, then there will be an undesirable amount of delay before the imitating agent will be successful.
	Furthermore, algorithms that require post-demonstration interaction typically require it again and again for each newly-demonstrated task, which could result in even more delay.
	Beyond delay, environment interaction can also be risky.
	For example, when training autonomous vehicles, operating on city streets while learning might endanger lives or lead to costly damage due to crashes.
	Therefore, we desire an algorithm for which environment interactions can be performed as a pre-processing step - perhaps in a safer environment - and where the information learned from those interactions can be re-used for a variety of demonstrated tasks.
	

	In this paper, we propose a new imitation learning algorithm called \textit{behavioral cloning from observation (BCO)}.
	BCO simultaneously addresses both of the issues discussed above, i.e., it provides reasonable imitation policies almost immediately upon observing state-trajectory-only demonstrations.
	First, it calls for the agent to learn a task-independent, inverse dynamics model in a pre-demonstration, exploratory phase.
	Then, upon observation of a demonstration without action information, BCO uses the learned model to infer the missing actions.
	Finally, BCO uses the demonstration and the inferred actions to find a policy via behavioral cloning.
	If post-demonstration environment interaction is allowed, BCO additionally specifies an iterative scheme where the agent uses the extra interaction time in order to learn a better model and improve its imitation policy.
	This iterative scheme therefore provides a tradeoff between imitation performance and post-demonstration environment interaction.

	\section{Related Work}\label{related-work}
	BCO is related to both imitation learning and model-based learning.
	We begin with a review of imitation learning approaches,  which typically fall under one of two broad categories: \textit{behavioral cloning (BC)} and \textit{inverse reinforcement learning (IRL)}.
	
	Behavioral cloning \cite{bain1999framework,ross2011reduction,daftry2016learning} is one of the main methods to approach an imitation learning problem.
	The agent receives as training data both the encountered states and actions of the demonstrator, and then uses a classifier or regressor to replicate the expert's policy \cite{ross2010efficient}.
	This method is powerful in the sense that it is capable of imitating the demonstrator immediately without having to interact with the environment.
	Accordingly, BC has been used in a variety of applications.
	For instance, it has been used to train a quadrotor to fly down a forest trail \cite{giusti2016machine}.
	There, the training data is the pictures of the forest trail labeled with the actions that the demonstrating quadrotor used, and the policy is modeled as a convolutional neural network classifier.
	In the end, the quadrotor manages to fly down the trail successfully.
	BC has also been used in autonomous driving \cite{bojarski2016end}.
	The training data is acquired from a human demonstrator, and a convolutional neural network is trained to map raw pixels from a single front-facing camera directly to steering commands.
	After training, the vehicle is capable of driving in traffic on local roads. 
	BC has also been successfully used to teach manipulator robots complex, multi-step, real-world tasks using kinesthetic demonstration \cite{niekum2015learning}.
	While behavioral cloning is powerful, it is also only applicable in scenarios where the demonstrator's action sequences are available.
	However, when humans want to imitate each other's actions, they do not have access to the internal control signals the demonstrator used.
	Instead, they only see the effects of those actions.
	In our setting, we wish to perform imitation in these scenarios, and so we cannot apply BC techniques as they are typically formulated.
	
	Inverse reinforcement learning is a second category of imitation learning.
	IRL techniques seek to learn a cost function that has the minimum value for the demonstrated actions.
	The learned cost function is then used in combination with RL methods to find an imitation policy.
	Like BC techniques, IRL methods usually assume that state-action pairs are available \cite{finn2016guided,ho2016generative,ho2016model}, and also that the reward is a function of both states and actions.
	An exception is the work of Liu {\em et al.} [2017]. 
	In this work, it is assumed that both demonstrator and imitator are capable of following a trajectory at the exact same pace to perform a task, and the IRL method defines the reward signal to be the proximity of the imitator and demonstrator's encoded state features at each time step.
	As a result, the reward signal can only be generated after the demonstration is made available, after which reinforcement learning and environment interaction must be completed in order to find a good policy.
	In our work, we wish to minimize the amount of environment interaction necessary after the demonstration is provided, and so we seek an alternative to IRL.
	
	BCO is also related to the model-based learning literature in that it makes use of learned models of the environment.
	In general, model-based methods have major advantages over those that are model-free.
	First, they are more sample-efficient \cite{chebotar2017combining}, i.e., they do not require as many environment interactions as model-free methods.
	Second, the learned models can be transferred across tasks \cite{taylor2008transferring}.
	Typical model-learning techniques focus on obtaining an estimate of the transition dynamics model, i.e., a mapping from current state and action to the next state.
	In our work, on the other hand, we want the agent to learn a model of the environment that will help us infer missing actions, and therefore BCO learns a slightly-different inverse dynamics model, i.e., a mapping from state transitions to the actions \cite{hanna2017grounded}.
	
	There has also been recent work done where inverse models have been used to perform imitation learning in the absence of action information. Niekum et al [2015a] present such a technique for situations in which the kinematic equations are known. Nair et al [2017] propose a technique that first learns an inverse dynamics model and then use that model to estimate the missing action information at each time step from a single demonstration. The method we develop here, on the other hand, both does not assume prior knowledge of the inverse model and is capable of generalizing in cases when mulitiple demonstrations are available.

	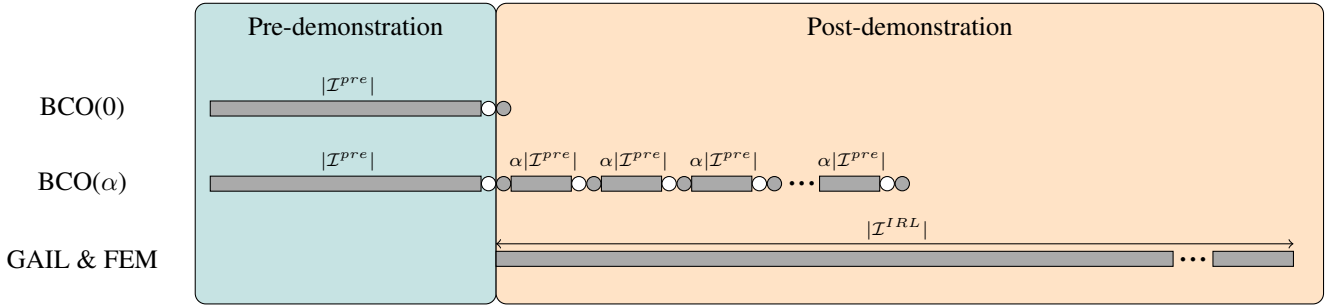
\begin{figure*}[!ht]
\begin{center}
\begin{tikzpicture}

\node [text=black, align=center] at (1.5,2.6) {BCO(0)};

\node [text=black, align=center] at (1.5,1.6) {BCO($\alpha$)};

\node [text=black, align=center] at (1.5,0.6) {GAIL \& FEM};

\draw [fill={rgb:blue,1;green,1;white,7}, rounded corners] (3,0) rectangle (7,4);

\node [text=black, align=center] at (5,3.7) {Pre-demonstration};

\draw [fill={rgb:yellow,1;red,1;white,7}, rounded corners] (7,0) rectangle (18,4);

\node [text=black, align=center] at (12.5,3.7) {Post-demonstration};

\draw [fill={rgb:black,1;white,2}] (3.2,2.5) rectangle (6.8,2.7);
\draw [fill=white] (6.9,2.6) circle [radius=0.095];
\node [text=black, align=center] at (5,2.9) {\scalefont{0.7} $|\I^{pre}|$};
\draw [fill={rgb:black,1;white,2}] (3.2,1.5) rectangle (6.8,1.7);
\draw [fill=white] (6.9,1.6) circle [radius=0.095];
\node [text=black, align=center] at (5,1.9) {\scalefont{0.7} $|\I^{pre}|$};

\draw [fill={rgb:black,1;white,2}] (7.1,2.6) circle [radius=0.095];
\draw [fill={rgb:black,1;white,2}] (7.1,1.6) circle [radius=0.095];
\draw [fill={rgb:black,1;white,2}] (7.2,1.5) rectangle (8,1.7);
\node [text=black, align=center] at (7.6,1.9) {\scalefont{0.7} $\alpha |\I^{pre}|$};
\draw [fill=white] (8.1,1.6) circle [radius=0.095];
\draw [fill={rgb:black,1;white,2}] (8.3,1.6) circle [radius=0.095];
\draw [fill={rgb:black,1;white,2}] (8.4,1.5) rectangle (9.2,1.7);
\node [text=black, align=center] at (8.8,1.9) {\scalefont{0.7} $\alpha |\I^{pre}|$};
\draw [fill=white] (9.3,1.6) circle [radius=0.095];
\draw [fill={rgb:black,1;white,2}] (9.5,1.6) circle [radius=0.095];
\draw [fill={rgb:black,1;white,2}] (9.6,1.5) rectangle (10.4,1.7);
\node [text=black, align=center] at (10,1.9) {\scalefont{0.7} $\alpha |\I^{pre}|$};
\draw [fill=white] (10.5,1.6) circle [radius=0.095];
\draw [fill={rgb:black,1;white,2}] (10.7,1.6) circle [radius=0.095];
\node [text=black, align=center] at (11,1.6) {\scalefont{1.5} ...};
\draw [fill={rgb:black,1;white,2}] (11.3,1.5) rectangle (12.1,1.7);
\node [text=black, align=center] at (11.7,1.9) {\scalefont{0.7} $\alpha |\I^{pre}|$};
\draw [fill=white] (12.2,1.6) circle [radius=0.095];
\draw [fill={rgb:black,1;white,2}] (12.4,1.6) circle [radius=0.095];
\draw [fill={rgb:black,1;white,2}] (7,.5) rectangle (16,.7);
\node [text=black, align=center] at (16.2,.6) {\scalefont{1.5} ...};
\draw [fill={rgb:black,1;white,2}] (16.53,0.5) rectangle (17.6,.7);
\draw [<->] (7,.8) -- (17.6,.8);
\node [text=black, align=center] at (12.3,1) {\scalefont{0.7} $|\I^{IRL}|$};

\end{tikzpicture}
\caption{Learning timelines for BCO($0$), BCO($\alpha$), and the IRL methods we compare against in this paper. The horizontal axis represents time, gray rectangles mark when each technique requires environment interactions. For BCO, the white and gray circles denote the inverse model and policy learning steps, respectively. For BCO, $\alpha |\I^{pre}|$ is the number of post-demonstration environment interactions performed before each model- and policy-improvement step and for IRL methods, $|\I^{IRL}|$ represents the total number of interactions. }
			\label{fig:pre-post}
\end{center}
\end{figure*}	
	
	\section{Problem Formulation}\label{formulation}
	We consider agents acting within the broad framework of Markov decision processes (MDPs).
	We denote a MDP using the 5-tuple $M=\{S, A, T, r, \gamma \}$, where $S$ is the agent's state space, $A$ is its action space, $T^{s_i a}_{s_{i+1}} = P(s_{i+1}|s_i,a)$ is a function denoting the probability of the agent transitioning from state $s_i$ to $s_{i+1}$ after taking action $a$, $r: S \times A \rightarrow \mathbb{R}$ is a function specifying the immediate reward that the agent receives for taking a specific action in a given state, and $\gamma$ is a discount factor.
	In this framework, agent behavior can be specified by a policy, $\pi: S \rightarrow A$, which specifies the action (or distribution over actions) that the agent should use when in a particular state.
	We denote the set of state transitions experienced by an agent during a particular execution of a policy ${\pi}$ by $\Tau_{\pi} = \{(s_i, s_{i+1})\}$.
	
	In the context of these transitions, we will be interested in the {\em inverse dynamics model}, $\Mu^{s_i s_{i+1}}_{a} = P(a|s_i,s_{i+1})$, which is the probability of having taken action $a$ given that the agent transitioned from state $s_i$ to $s_{i+1}$.
	Moreover, we specifically seek task-independent models.
	We assume that some of the state features are specifically related to the task and others specifically to the agent, i.e., a given state $s$ can be partitioned into an {\em agent-specific state}, $s^a$, and a {\em task-specific state}, $s^t$, which are members of sets $S^a$ and $S^t$, respectively (i.e., $S = S^a \times S^t$) \cite{konidaris2006framework,gupta2017learning}. 
	Using this partitioning, we define the {\em agent-specific inverse dynamics model} to be a function $\Mu_\theta: S^a \times S^a \rightarrow p(A)$ that maps a pair of agent-specific state transitions, $(s_i^a, s_{i+1}^a) \in \Tau^a_{\pi}$, to a distribution of agent actions that is likely to have given rise to that transition.
	
	Imitation learning is typically defined in the context of a MDP {\em without} an explicitly-defined reward function, i.e., $M\setminus r$.
	The learning problem is for an agent to determine an {\em imitation policy}, $\pi: S \rightarrow A$ that the agent may use in order to behave like the expert, using a provided set of expert demonstrations $\{ \xi_1, \xi_2, ... \}$ in which each $\xi$ is a demonstrated state-action trajectory $\{(s_0,a_0), (s_1,a_1),...,(s_N,a_N)\}$. Therefore, in this setting, the agent must have access to the demonstrator's actions.
	If the imitator is unable to observe the action sequences used by the demonstrator, the resulting imitation learning problem has recently been referred to as {\em imitation from observation} \cite{liu2017imitation}.
	In this setting, one seeks to find an imitation policy from a set of state-only demonstrations $D = \{ \zeta_1, \zeta_2, ... \}$ in which each $\zeta$ is a state-only trajectory $\{s_0, s_1, ...,s_N\}$.

	The specific problem that we are interested in is imitation from observation under a constrained number of environment interactions. 
	By {\em environment interactions} we mean time steps for which we require our agent to gather new data by executing an action in its environment and observing the state outcome.
	We are concerned here in particular with the cost of the learning process, in terms of the number of environment interactions, both before and after the expert demonstrations are provided.
	Pre- and post-demonstration environment interactions are represented by $\I^{pre} $ and $\I^{post}$, respectively, to denote sets of interactions $(s_i,a_i,s_{i+1})$ that must be executed by a learner before and after a demonstration becomes available.
	In this context, we are concerned here with the following specific goal: {\em given a set of state-only demonstration trajectories, $D$, find a good imitation policy using a minimal number of post-demonstration environment interactions, i.e., $|\I^{post}|$.}
	
	In pursuit of this goal, we propose a new algorithm for imitation learning that can operate both in the absence of demonstrator action information and while requiring no or very few post-demonstration environment interactions.
	Our framework consists of two components, each of which considers a separate part of this problem.
	The first of these components considers the problem of learning an agent-specific inverse dynamics model, and the second one considers the problem of learning an imitation policy from a set of demonstration trajectories.

	\section{Behavioral Cloning from Observation }\label{BCO}

	We now describe our imitation learning algorithm, BCO, which combines inverse dynamics model learning with learning an imitation policy.
	We are motivated by the fact that humans have access to a large amount of prior experience about themselves, and so we aim to also provide an autonomous agent with this same prior knowledge.
	To do so, before any demonstration information is observed, we allow the agent to learn its own agent-specific inverse dynamics model.
	Then, given state-only demonstration information, we use this learned model to infer the expert's missing action information.
	Once these actions have been inferred, the agent performs imitation learning via a modified version of behavioral cloning (Figure \ref{fig:framework}). The pseudo-code of the algorithm is given in Algorithm \ref{bco}.

	\subsection{Inverse Dynamics Model Learning}
	\label{sec:inversemodellearning}

	In order to infer missing action information, we first allow the agent to acquire prior experience in the form of an agent-specific inverse dynamics model.
	In order to do so, we let the agent perform an exploration policy, $\pi$.
	In this work, we let $\pi$ be a random policy (Algorithm \ref{bco}, Line 2).
	While executing this policy, the agent performs some number of interactions with the environment, i.e., $\I^{pre}$. 
	Because we seek an agent-specific inverse dynamics model as described in Section \ref{formulation}, we extract the agent-specific part of the states in $\I^{pre}$ and store them as
	$\Tau^a_{\pi_e}=\{(s_i^a,s_{i+1}^a)\}$, and their associated actions, $\Alpha_{\pi_e}=\{a_i\}$ (Algorithm \ref{bco}, Lines 5-8).
	Given this information, the problem of learning an agent-specific inverse dynamics model is that of finding the parameter $\theta$ for which $\Mu_\theta$ best describes the observed transitions.
	We formulate this problem as one of maximum-likelihood estimation, i.e., we seek $\theta^*$ as
	
	\begin{equation}
	\theta^* = \text{arg} \; \text{max}_\theta \; \Pi_{i=0}^{|\I^{pre}|} \; p_{\theta}(a_i \; | s_i^a, s_{i+1}^a) ,
	\label{eq:modelprogram}
	\end{equation}
	
	\noindent
	where $p_\theta$ is the conditional distribution over actions induced by $\Mu_\theta$ given a specific state transition. 
	Any number of supervised learning techniques, denoted as ``modelLearning" in Algorithm \ref{bco}, may be used to solve (\ref{eq:modelprogram}).
	
	Some details regarding particular choices made in this paper: For domains with a continuous action space, we assume a Gaussian distribution over each action dimension and our model estimates the individual means and standard deviations.
	We use a neural network for $\Mu_\theta$, where the network receives a state transition as input and outputs the mean for each action dimension.
	The standard deviation is also learned for each dimension, but it is computed independently of the state transitions.
	In order to train this network (i.e., to find $\theta^*$ in (\ref{eq:modelprogram})), we use  the Adam variant \cite{kingma2014adam} of stochastic gradient decent.
	Intuitively, the gradient for each sample is computed by finding a change in $\theta$ that would increase the probability of $a_i$ with respect to the distribution specified by $\Mu_{\theta}(s_i,s_{i+1})$.
	When the action space is discrete, we again use a neural network for $\Mu_{\theta}$, where the network computes the probability of taking each action via a softmax function.
	
	\subsection{Behavioral Cloning}

	Our overarching problem is that of finding a good imitation policy from a set of state-only demonstration trajectories, $D_{demo} = \{ \zeta_1, \zeta_2, \ldots \}$ where each $\zeta$ is a trajectory $\{s_0, s_1, \ldots, s_N\}$.
	Note that, although the inverse dynamics model is learned using a set of agent-generated data in Section \ref{sec:inversemodellearning}, the data used there is not utilized in this step.
	In order to use the learned agent-specific inverse dynamics model, we first extract the agent-specific part of the demonstrated state sequences
	and then form the set of demonstrated agent-specific state transitions $\Tau^a_{demo}$ (Algorithm \ref{bco}, Line 10).
	Next, for each transition $( s_{i}^{a} , s_{i+1 }^{a} ) \in \Tau^a_{demo}$, the algorithm computes the model-predicted distribution over demonstrator actions, $\Mu_{\theta^*}(s_{i}^{a} , s_{i+1}^{a})$ and uses the maximum-likelihood action as the inferred action, $\tilde{a}_i$, which is placed in a set $\tilde{\Alpha}_{demo}$ (Algorithm \ref{bco}, Line 11).
	Using these inferred actions, we then build the set of complete state-action pairs $\{(s_i, \tilde{a}_i)\}$.
	
	With this new set of state-action pairs, we may now seek the imitation policy $\pi_{\phi}$.
	We cast this problem as one of behavioral cloning, i.e., given a set of state-action tuples $\{ (s_i,\tilde{a}_i) \}$, the problem of learning an imitation policy becomes that of finding the parameter $\phi$ for which $\pi_{\phi}$ best matches this set of provided state-action pairs (Algorithm \ref{bco}, Line 12).
	We find this parameter using maximum-likelihood estimation, i.e., we seek $\phi^*$ as
	
	\begin{equation}
	\phi^* = \text{arg} \; \text{max}_{\phi} \; \Pi_{i=0}^N \pi_{\phi}(\tilde{a}_i \; | \; s_i) \; .
	\label{eq:policyprogram}
	\end{equation}
	
	Some details regarding particular choices made in this paper: For continuous action spaces, we assume our policy to be Gaussian over each action dimension, and, for discrete actions spaces we use a softmax function to represent the probability of selecting each value.
	We let $\pi$ be a neural network that receives as input a state and outputs either the Gaussian distribution parameters or the action probabilities for continuous or discrete action spaces, respectively.
	We then solve for $\phi^*$ in (\ref{eq:policyprogram}) using Adam SGD, where the intuitive view of the gradient is that it seeks to find changes in $\phi$ that increase the probability of each inferred demonstrator action, $\tilde{a}_i$, in the imitation policy's distribution $\pi_{\phi}(\cdot \; | \; s_i)$.
	
	\begin{algorithm}[t!]
		\caption{BCO($\alpha$)}\label{bco}
		\begin{algorithmic}[1]
			\State Initialize the model $\Mu_\theta$ as random approximator
			\State Set $\pi_\phi$ to be a random policy
			
			\State Set $I=|\I^{pre}|$
			
			\While {policy improvement}
			\For{time-step t=1 to $I$}
			\State Generate samples $(s^a_t, s^a_{t+1})$ and $a_t$ using $\pi_\phi$
			\State Append samples $\Tau^a_{\pi_\phi} \gets (s^a_t, s^a_{t+1})$, $\Alpha_{\pi_\phi} \gets a_t$ 
			\EndFor

			\State Improve $\Mu_\theta$ by modelLearning($\Tau^a_{\pi_\phi}$, $\Alpha_{\pi_\phi}$)
			\State Generate set of agent-specific state transitions $\Tau^a_{demo}$ \-\ \-\ \-\ \-\ \-\ \-\ from the demonstrated state trajectories $D_{demo}$
			\State Use $\Mu_{\theta}$ with $\Tau^a_{demo}$ to approximate $\tilde{\Alpha}_{demo}$
			
			\State Improve $\pi_\phi$ by behavioralCloning($S_{demo}, \tilde{\Alpha}_{demo}$)
			\State Set $I=\alpha |\I^{pre}|$
			\EndWhile
			
		\end{algorithmic}
	\end{algorithm}

\begin{figure*}
	\begin{subfigure}{.24\textwidth}
		\centering
		\includegraphics[width=.9\linewidth, height=3cm]{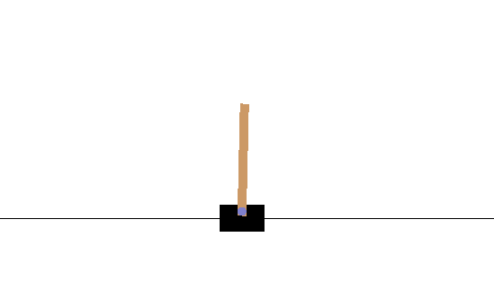}
		\caption{CartPole}
		\label{fig:cartpole}
	\end{subfigure}
	\begin{subfigure}{.24\textwidth}
		\centering
		\includegraphics[width=.9\linewidth, height=3cm]{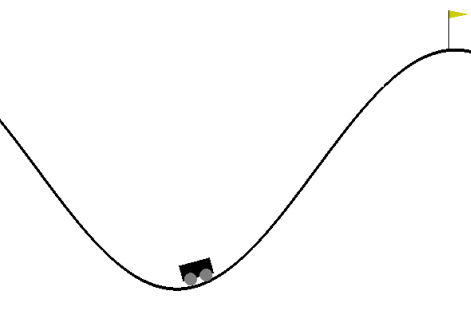}
		\caption{MountainCar}
		\label{fig:mountaincar}
	\end{subfigure}
	\begin{subfigure}{.24\textwidth}
		\centering
		\includegraphics[width=.9\linewidth, height=3cm]{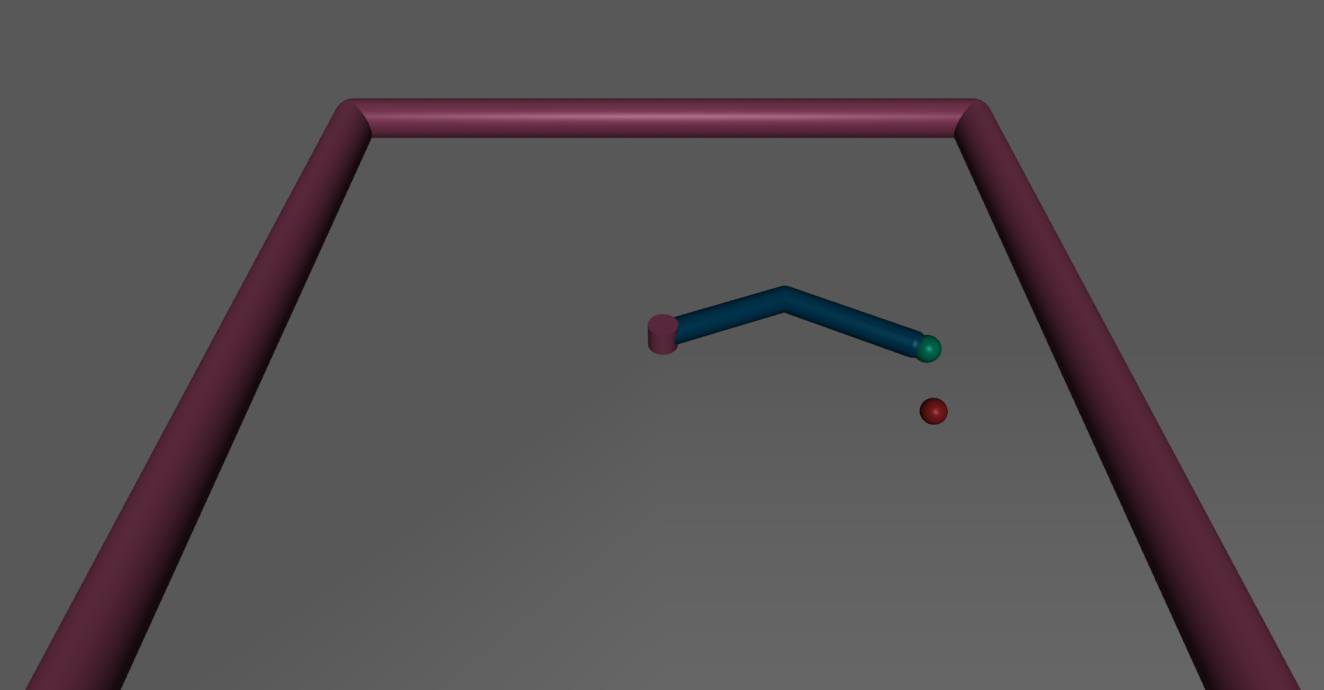}
		\caption{Reacher}
		\label{fig:reacher}
	\end{subfigure}
	\begin{subfigure}{.24\textwidth}
		\centering
		\includegraphics[width=.9\linewidth, height=3cm]{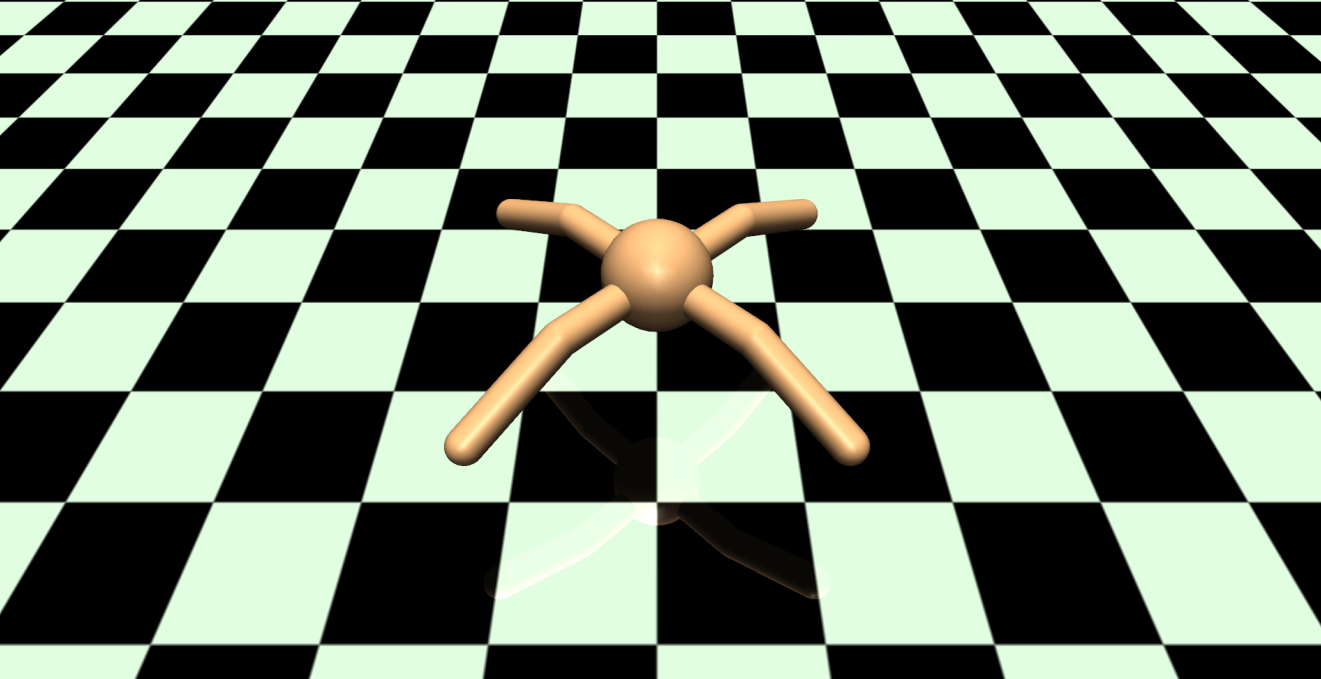}
		\caption{Ant}
		\label{fig:ant}
	\end{subfigure}%
	\caption{Representative screenshots of the domains considered in this paper.}
	\label{fig:domains}
\end{figure*}
	
	\subsection{Model Improvement}
	
	The techniques described above form the building blocks of BCO.
	If one is willing to tolerate post-demonstration environment interaction, a modified version of our algorithm can further improve both the learned model and the resulting imitation policy.
	This modified algorithm proceeds as follows.
	After the behavioral cloning step, the agent executes the imitation policy in the environment for a short period of time. 
	Then, the newly-observed state-action sequences are used to update the model, and, accordingly, the imitation policy itself.
	The above procedure is repeated until there is no more improvement in the imitation policy.
	We call this modified version of our algorithm BCO($\alpha$), where $\alpha$ is a user-specified parameter that is used to control the number of post-demonstration environment interactions at each iteration, $M$, according to $M = \alpha |\I^{pre}|$.
	The total number of post-demonstration interactions required by BCO($\alpha$) can be calculated as $|\I^{pre}| = TM = T \alpha |\I^{pre}|$, where $T$ is the total number of model-improvement iterations required by BCO($\alpha$). 
	Using a nonzero $\alpha$, the model is able to leverage post-demonstration environment interaction in order to more accurately estimate the actions taken by the demonstrator, and therefore improve its learned imitation policy.
	If one has a fixed budget for post-demonstration interactions, one could consider terminating the model-improvement iterations early, i.e., specify both $\alpha$ and $T$.

	\section{Implementation and Experimental Results}\label{experiments}
	
	We evaluated BCO($\alpha$) in several domains available in OpenAI Gym \cite{brockman2016openai}. Continuous tasks are simulated by MuJoCo \cite{todorov2012mujoco}.
	These domains have different levels of difficulty, as measured by the complexity of the dynamics and the size and continuity of the state and action spaces.
	Ordered from easy to hard, the domains we considered are: {\bf CartPole}, {\bf MountainCar}, {\bf Reacher}, and {\bf Ant-v1}.
	Each of these domains has predefined state features, actions and reward functions in the simulator. Any RL algorithm could be used with this reward function to generate an expert policy.
	Since trust region policy optimization (TRPO) \cite{schulman2015trust} has shown promising performance in simulation \cite{liu2017imitation}, we generated our demonstrations using agents trained with this method.
	
	We evaluated our algorithm in two senses.
	First, with respect to the number of environment interactions required to attain a certain performance. In a real-world environment, interactions can be expensive which makes it a very important criterion.
	The second way in which we evaluate our algorithm is with respect to data efficiency, i.e., the imitator's task performance as a function of the amount of available demonstration data.
	In general, demonstration data is scarce, and so making the best use of it is very important.

	\begin{figure*}[!ht]
		\centering
		\includegraphics[scale=.55]{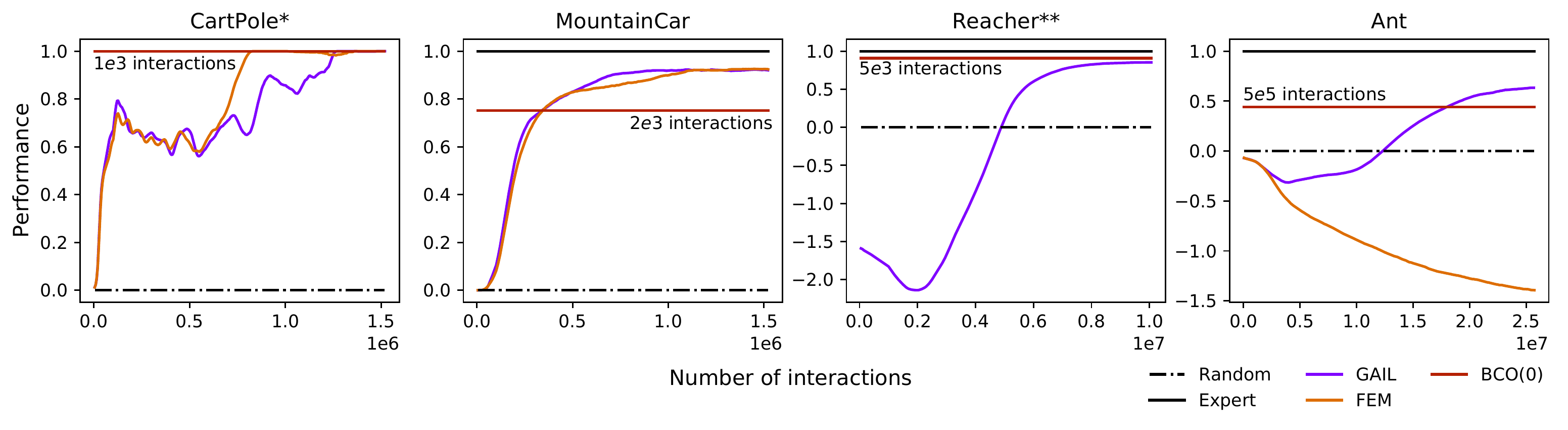}
		\caption{Performance of each technique with respect to the number of post-demonstration interactions. For each domain, ten demonstrated trajectories were considered. BCO(0) is depicted as a horizontal line since all environment interactions happen before the demonstration is provided. Performance values are scaled such that performance of a random policy is zero and the performance of the expert is one. Note that GAIL and FEM have access to demonstration action information whereas BCO does not. *The BCO line is not visible for the CartPole domain because BCO has the same performance as the expert. **FEM is not shown for the Reacher domain because its performance is much worse than the other techniques.}
		\label{fig:interaction}. 
	\end{figure*}
	
	\begin{figure*}[!ht]
		\centering
		\includegraphics[scale=.56]{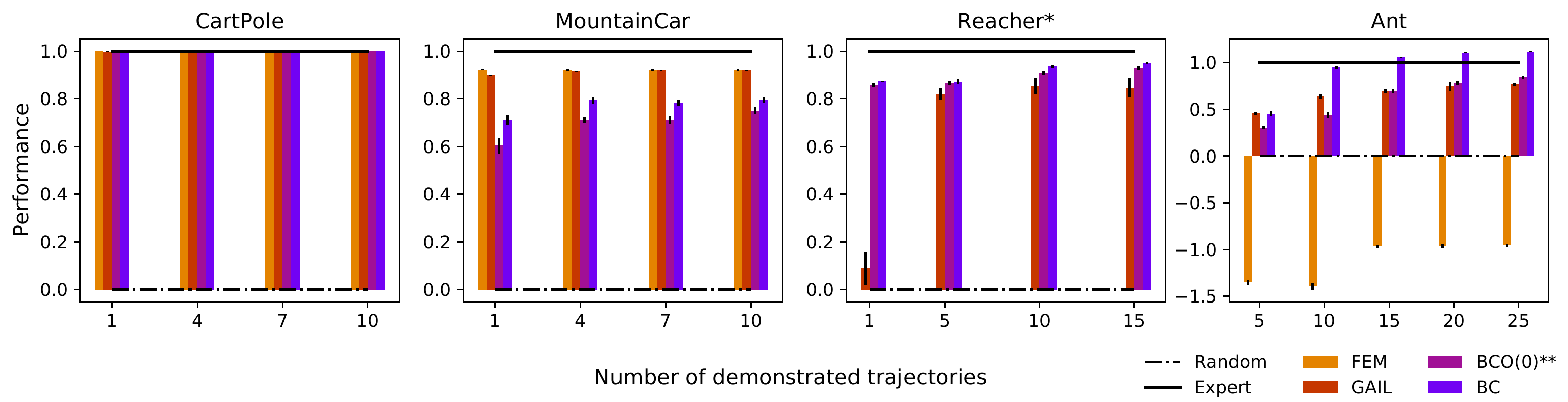}
		\caption{Performance of imitation agents with respect to the number of available demonstration trajectories. Rectangular bars and error bars represent the mean return and the standard error, respectively, as measured over 5000 trajectories. Returns have been scaled such that the performance of a random policy and the demonstrating agent are zero and one, respectively. *Note that FEM is not shown for the Reacher domain because its performance is much worse than the others. 
			**Note that BC, GAIL, and FEM all have access to demonstration action information whereas {\em BCO(0) does not.}}
		\label{fig:performance}
	\end{figure*}

	We compared BCO($\alpha$) to the following methods:
	\begin{enumerate}
		\item {\bf Behavioral Cloning (BC):} This method applies supervised learning over state-action pairs provided by the demonstrator.
		\item {\bf Feature Expectation Matching (FEM) \cite{ho2016model}:} A modified version of the approach presented by Abbeel and Ng [2004]. It uses trust region policy optimization with a linear cost function in order to train train neural network policies.
		\item {\bf General Adversarial Imitation Learning (GAIL) \cite{ho2016generative}:} A state-of-the-art in IRL. It uses a specific class of cost functions which allows for the use of generative adversarial networks in order to do apprenticeship learning.
	\end{enumerate}
	Note in particular that {\em \bf our method is the only method that does not have access to the demonstrator's actions}.
	However, as our results will show, BCO($\alpha$) can still achieve comparable performance to these other techniques, and do so while requiring far fewer environment interactions.

	\subsection{Training Details and Results}
	Because both BC and BCO($\alpha$) rely on supervised learning methods, we use only 70\% of the available data for training and use the rest for validation.
	We stop training when the error on the 30\% validation data starts to increase.
	For the other methods, all available data was used in the training process.
	We will now discuss the architecture, details for each domain.
	
	\begin{itemize}
		\item {\bf CartPole:} The goal is to keep the pole vertically upward as long as possible (Figure \ref{fig:cartpole}). This domain has a discrete action space.
		In this domain, we considered linear models over the pre-defined state features for both the inverse dynamics model and the imitation policy and we only used $M=1000$ interactions to learn the dynamics.
		
		\item {\bf  MountainCar:} The goal is to have the car reach the target point (Figure \ref{fig:mountaincar}). This domain has a discrete action space.
		In this domain, the data set for learning the inverse dynamics model is acquired by letting the agent to explore its action space for $M=2000$ time steps.
		For both the imitation policy and inverse dynamics model, we used neural networks with two hidden layers, 8 nodes each, and leaky rectified linear activation functions (LReLU).
		
		\item {\bf Reacher:} This domain has a continuous action space. The goal is to have the fingertip of the arm reach the target point whose position changes in every episode (Figure \ref{fig:reacher}). Therefore, in this domain, we can partition the state-space to agent-specific features (i.e., those only related to the arm) and task-specific features (i.e., those related to the position of the target). A neural network architecture with two hidden layers of 100 LReLU nodes are used with $M=5000$ agent-specific state transition-action pairs in order to learn the dynamics and then this model is used to learn a policy which also has two layers but with 32 LReLU nodes.
		
		\item {\bf  Ant:} The goal to have the ant to run as fast as possible(Figure \ref{fig:ant}). This domain has a continuous action space.
		This is the most complex domain considered in this work.
		The state and action space are 111 and 8 dimensional, respectively. The number of interactions needed to learn the dynamics was $M=5e5$ and the architectures for inverse dynamics learning and the policy are similar to those we used in {\bf Reacher}.
	\end{itemize}
	
	\subsection{Discussion}
	Each experiment was executed twenty times, and all results presented here are the average values over these twenty runs.
	We selected twenty trials because we empirically observed very small standard error bars in our results.
	This is likely a reflection of the relatively low variance in the expert demonstrations.
	
	In our first experiment, we compare the number of environment interactions needed for BCO(0) with the number required by other methods (Figure \ref{fig:interaction}.)
	We can clearly see how imitation performance improves as the agent is able to interact more with the environment.
	In the case of BCO(0), the interactions with the environment happen before the policy-learning process starts, and so we represent its performance with a horizontal line.
	The height of the line indicates the performance, and we display the number of pre-demonstration environment interactions it required next to it.
	The random and expert policies also do not benefit from post-demonstration environment interaction, and so they are also shown using horizontal lines.
	From these plots, we can see that it takes at least 40 times more interactions required by GAIL or FEM to gain the same performance as BCO(0).

	\begin{figure*}[!ht]
		\centering
		\includegraphics[scale=.6]{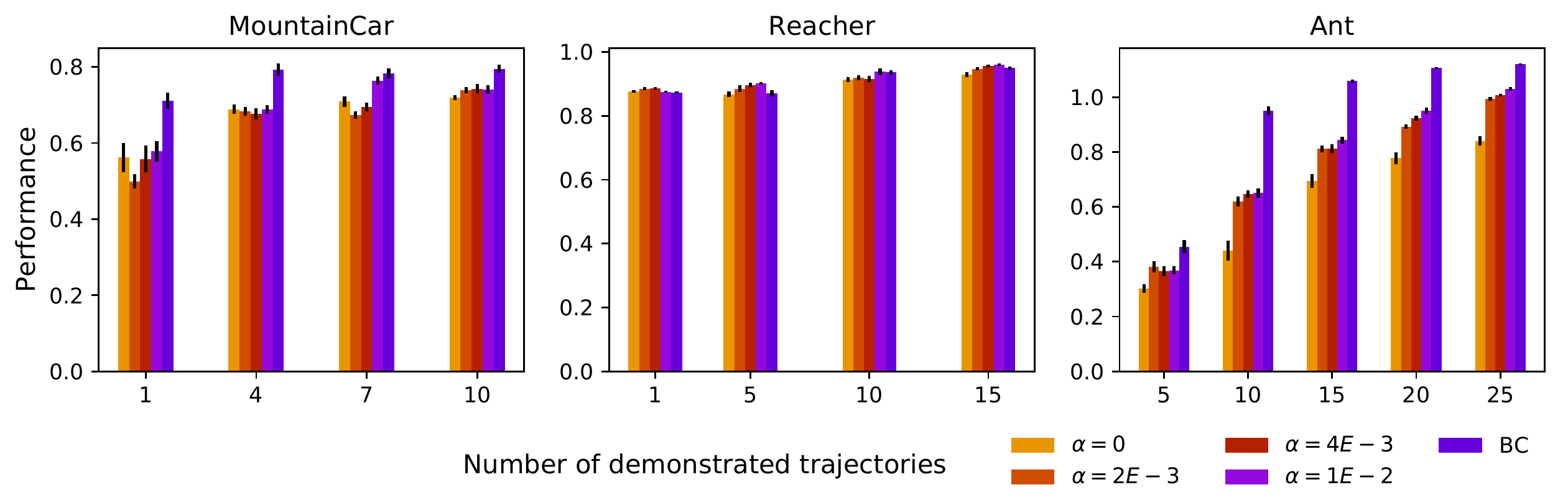}
		\caption{The performance of BC and several BCO($\alpha$) techniques (varying $\alpha$) with respect to the number of demonstrated trajectories provided. Rectangular bars and error bars represent the mean return and the standard error, respectively, as measured over 5000 trajectories. By increasing $\alpha$, more post-demonstration environment interactions are allowed to occur, which increases the performance of the imitation policy. Note that BC has access to demonstration action information whereas BCO does not. Also note that the number of trajectories required for learning a fairly good policy is very small. Each demonstrated trajectory has 5, 50, and 50 transitions for each domain from left to right, respectively. Note that we did not demonstrate the results for CartPole because the results were equally perfect regardless of the value of $\alpha$.}
		\label{fig:alpha}
	\end{figure*}

\begin{table*}[ht]
\centering
\begin{tabular}{|M{.9cm}|M{1.1cm}|M{1.1cm}|M{1.1cm}|M{.9cm}|M{1.1cm}|M{1.1cm}|M{1.1cm}|M{.9cm}|M{1.1cm}|M{1.1cm}|M{1.1cm}|N}
\hline
 \multicolumn{4}{|c|}{\bf{MountainCar (pre-demo \boldmath$=2E3$)}}& \multicolumn{4}{|c|}{\bf{Reacher (pre-demo \boldmath$=5E3$)}}& \multicolumn{4}{|c|}{\bf{Ant (pre-demo \boldmath$=5E5$)}} &\\[6pt]
\hline
 
 $\bf{d}$ \boldmath$\bf{\backslash}$ $\bf{\alpha}$ & \boldmath$2E-3$ & \boldmath$4E-3$ & \boldmath$1E-2$ & $\bf{d}$ \boldmath$\bf{\backslash}$ $\bf{\alpha}$ & \boldmath$2E-3$ & \boldmath$4E-3$ & \boldmath$1E-2$ & $\bf{d}$ \boldmath$\bf{\backslash}$ $\bf{\alpha}$ & \boldmath$2E-3$ & \boldmath$4E-3$ & \boldmath$1E-2$ &\\[6pt]
\hline
\boldmath$1$ & $6825$ & $23475$ & $28950$ & \boldmath$1$ & $210052$ & $358736$ & $912368$ & \boldmath$5$ & $602500$ & $1270000$ & $3362500$ &\\[6pt]
\hline
\boldmath$4$ & $8387$ & $12000$ & $31200$ & \boldmath$5$ & $270500$ & $486578$ & $1837500$ & \boldmath$10$ & $940000$ & $2075000$ & $5000000$ &\\[6pt]
\hline
\boldmath$7$ & $6300$ & $23100$ & $122200$ & \boldmath$10$ & $221421$ & $569736$ & $1055921$ & \boldmath$15$ & $1387500$ & $2855000$ & $7325000$ &\\[6pt]
\hline
\boldmath$10$ & $45462$ & $61450$ & $88600$ & \boldmath$15$ & $509289$ & $852210$ & $1859210$ & \boldmath$20$ & $1925000$ & $4055000$ & $9687500$ &\\[6pt]
\hline
\end{tabular}
		\caption{This is an extension to Figure \ref{fig:alpha} which provides the number of post-demonstration interactions for each case. First vertical column for each domain ($d$) shows the number of demonstrated trajectories. As an example, for the MountainCar domain with $1$ demonstrated trajectory and $\alpha=2E-3$, the average number of post-demonstration interactions is 6825. We can also see at the top of the table that the number of pre-demonstration interactions is $2E3$ so the overall number of interactions would become $8825$. It can be seen that almost always by increasing $\alpha$ or the number of demonstrated trajectories, the number of post-demonstration interactions increases. Also the overall number of interactions (combined pre- and post-demonstration interactions) in all the cases is far less than the overall number of interactions required by the other methods (FEM and GAIL).}
		\label{fig:interactions}
\end{table*}

	Now, we aim to compare the performance of our algorithm BCO($\alpha$) with the other algorithms. To do so, we use each algorithm to train the agents, and then calculate the final performance by computing the average return over $5000$ episodes. For comparison purposes, we scale these performance values such that the performance of the expert and a random policy are $1.0$ and $0.0$, respectively.
	This comparison is shown in Figures \ref{fig:performance}, and \ref{fig:alpha} where we have plotted the performance of each algorithm with respect to the number of available demonstrated trajectories. Figure \ref{fig:performance} shows the comparison between the performance of BCO(0) with all the other methods, and Figure \ref{fig:alpha} compares the performance of BCO($\alpha$) across different values of $\alpha$. In Figure \ref{fig:performance}, we can see that performance of our method is comparable with other methods even though our method is the only one without access to the actions. In the case of Reacher, the transferability of the learned inverse model is highlighted by the high performance of BCO. In the case of the Reacher and Ant domains, we can see that FEM performs poorly compared to others, perhaps because the rewards are not simple enough to be approximated by linear functions. In the CartPole domain, each of the methods performs as well as the expert and so all the lines are over each other. In the case of MountainCar, our performance is worse than other methods. Conversely, for Reacher, ours is more sample efficient than GAIL, i.e., with smaller number of demonstrations we get much better results. In the case of Ant, our method performs almost as good as GAIL.
	In Figure \ref{fig:alpha}, we can see that BCO's performance improves with larger $\alpha$ since the extra environment interactions allow it to make better estimates of the demonstrator's actions.

	\section{Conclusions}

	In this paper, we have presented BCO, an algorithm for performing imitation learning that requires neither access to demonstrator actions nor post-demonstration environment interaction.
	Our experimental results show that the resulting imitation policies perform favorably compared to those generated by existing imitation learning approaches that {\em do} require access to demonstrator actions.
	Moreover, BCO requires fewer post-demonstration environment interactions than these other techniques, meaning that a reasonable imitation policy can be executed with less delay.
	
	\section*{Acknowledgments}
	\thanks{Peter Stone serves on the Board of Directors of Cogitai, Inc.  The terms of this arrangement have been reviewed and approved by the University of Texas at Austin in accordance with its policy on objectivity in research.}
	
\bibliographystyle{named}
\bibliography{ijcai18}

\end{document}